\documentclass[11pt,a4paper]{article}
\usepackage[hyperref]{acl2019}
\usepackage{times}
\usepackage{latexsym}

\usepackage{url}

\usepackage{amssymb}
\usepackage{amsmath}
\usepackage{breqn}
\usepackage{graphicx}
\usepackage{graphics}
\usepackage{xspace}
\usepackage{booktabs}

\newcommand{\R}{\mathbb{R}}

\newcommand{\vocabin}{\mathcal{V}^{in}}
\newcommand{\vocabout}{\mathcal{V}^{out}}
\newcommand{\vocabattributes}{\mathcal{V}^{a}}
\newcommand{\vocabentities}{\mathcal{V}^{e}}

\newcommand{\atis}{\textsc{Atis}\xspace}
\newcommand{\geo}{\textsc{Geo}\xspace}

\newcommand{\webquestions}{\textsc{WebQuestionsSP}\xspace}
\newcommand{\spider}{\textsc{Spider}\xspace}

\newcommand\bertlarge{BERT$_{\small \textsc{LARGE}}$\xspace}

\aclfinalcopy 

\title{Generating Logical Forms from Graph Representations of Text and Entities}

\author{Peter Shaw$^1$, Philip Massey$^1$, Angelica Chen$^1$,
Francesco Piccinno$^2$, Yasemin Altun$^2$ \\
  $^1$Google \quad $^2$Google Research \\
  {\tt \{petershaw,pmassey,angelicachen,piccinno,altun\}@google.com}
}

\date{}

\begin{document}
\maketitle
\begin{abstract}

Structured information about entities is critical for many semantic parsing tasks. We present an approach that uses a Graph Neural Network (GNN) architecture to incorporate information about relevant entities and their relations during parsing. Combined with a decoder copy mechanism, this approach provides a conceptually simple mechanism to generate logical forms with entities. We demonstrate that this approach is competitive with the state-of-the-art across several tasks without pre-training, and outperforms existing approaches when combined with BERT pre-training.

\end{abstract}

\section{Introduction}

Semantic parsing maps natural language utterances into structured meaning representations.
The representation languages vary between tasks, but typically provide a precise, machine interpretable logical form suitable for applications such as question answering ~\cite{zelle1996learning, zettlemoyer2007online, liang2013learning, berant2013semantic}.
The logical forms typically consist of two types of symbols: a vocabulary of operators and domain-specific predicates or functions, and entities grounded to some knowledge base or domain.

Recent approaches to semantic parsing have cast it as a sequence-to-sequence task~\cite{dong2016language, jia2016data, ling2016latent},
employing methods similar to those developed for neural machine translation~\cite{bahdanau2014neural}, with strong results. However, special consideration is typically given to handling of entities. This is important to improve generalization and computational efficiency, as most tasks require handling entities unseen during training, and the set of unique entities can be large.

Some recent approaches have replaced surface forms of entities in the utterance with placeholders~\cite{dong2016language}. This requires a pre-processing step to completely disambiguate entities and replace their spans in the utterance. Additionally, for some tasks it may be beneficial to leverage relations between entities, multiple entity candidates per span, or entity candidates without a corresponding span in the utterance, while generating logical forms. 

Other approaches identify only types and surface forms of entities while constructing the logical form~\cite{jia2016data}, using a separate post-processing step to generate the final logical form with grounded entities. This ignores potentially useful knowledge about relevant entities.

Meanwhile, there has been considerable recent interest in Graph Neural Networks (GNNs) \cite{scarselli2009graph,li2016gated,kipf2016semi,gilmer2017neural,velivckovic2017graph} for effectively learning representations for graph structures. We propose a GNN architecture based on extending the self-attention mechanism of the Transformer~\cite{vaswani2017attention} to make use of relations between input elements. 

We present an application of this GNN architecture to semantic parsing, conditioning on a graph representation of the given natural language utterance and potentially relevant entities. This approach is capable of handling ambiguous and potentially conflicting entity candidates jointly with a natural language utterance, relaxing the need for completely disambiguating a set of linked entities before parsing. This graph formulation also enables us to incorporate knowledge about the relations between entities where available. Combined with a copy mechanism while decoding, this approach also provides a conceptually simple method for generating logical forms with grounded entities.

\begin{table*}[t!]

\begin{center}
\scalebox{0.9}{
\begin{tabular}{ll}
\toprule
\bf Dataset & \bf Example \\
\midrule
\geo & $\mathbf{x}:$ \textit{which states does the mississippi run through ?} \\
     & $\mathbf{y}:$ ~\texttt{answer}~(~\texttt{state}~(~\texttt{traverse\_1}(~\underline{riverid~(~mississippi~)}~)~)~) \\
\midrule

\atis & $\mathbf{x}:$ \textit{in denver what kind of ground transportation is there from the airport to downtown} \\

& $\mathbf{y}:$ (~\texttt{_lambda}~\texttt{\$0}~\texttt{e}~(~\texttt{_and}~(~ \texttt{_ground_transport}~\texttt{\$0}~)~(~\texttt{_to_city}\\
& ~~~~~~~~\texttt{\$0}~\underline{denver~:~\_ci}~)~(~\texttt{_from_airport}~\texttt{\$0}~\underline{den~:~\_ap}~)~)~)
\\
\midrule
\spider & $\mathbf{x}:$ \textit{how many games has each stadium held ?} \\

& $\mathbf{y}:$ 
~\texttt{SELECT}
~\texttt{T1}~.~\underline{id}~,
~\texttt{count}~(~$*$~)~\texttt{FROM}~\underline{stadium}~\texttt{AS}~\texttt{T1}~\texttt{JOIN}~\underline{game}~\texttt{AS} \\
 & ~~~~~~~~\texttt{T2}~~\texttt{ON}~\texttt{T1}~.~\underline{id}
 ~$=$~\texttt{T2}~.~\underline{stadium\_id}
 ~\texttt{GROUP}~\texttt{BY}~\texttt{T1}~.~\underline{id} \\
\bottomrule
\end{tabular}
}
\end{center}
\caption{Example input utterances, $\mathbf{x}$, and meaning representations, $\mathbf{y}$, with entities underlined.}
\label{semantic_parsing_tasks}
\end{table*}

We demonstrate the capability of the proposed architecture by achieving competitive results across 3 semantic parsing tasks. Further improvements are possible by incorporating a pre-trained BERT~\cite{devlin2018bert} encoder within the architecture.

\section{Task Formulation}

Our goal is to learn a model for semantic parsing from pairs of natural language utterances and structured meaning representations.
Let the natural language utterance be represented as a sequence $\mathbf{x} = (x_1, \ldots, x_{|\mathbf{x}|})$ of
$|\mathbf{x}|$ tokens, and the meaning representation be represented as a sequence $\mathbf{y} = (y_1, \ldots, y_{|\mathbf{y}|})$ of $|\mathbf{y}|$ elements.

The goal is to estimate $p(\mathbf{y} \mid \mathbf{x})$, the conditional probability of the meaning
representation $\mathbf{y}$ given utterance $\mathbf{x}$, which is augmented by a set of potentially relevant entities.

\paragraph{Input Utterance}
Each token $x_i \in \vocabin$ is from a vocabulary of input tokens.

\paragraph{Entity Candidates}

Given the input utterance $\mathbf{x}$, we retrieve a set, $\mathbf{e} = \{e_1, \ldots, e_{|\mathbf{e}|}\}$, of potentially relevant entity candidates, with $\mathbf{e} \subseteq \vocabentities$, where $\vocabentities$ is in the set of all entities for a given domain. We assume the availability of an entity candidate generator for each task to generate $\mathbf{e}$ given $\mathbf{x}$, with details given in \S~\ref{entity_candidate_generator}.

For each entity candidate, $e \in \vocabentities$, we require a set of task-specific attributes containing one or more elements from $\vocabattributes$.
These attributes can be NER types or other characteristics of the entity, such as ``city'' or ``river'' for some of the entities listed in Table~\ref{semantic_parsing_tasks}. 
Whereas $\vocabentities$ can be quite large for open domains, or even infinite if it includes sets such as the natural numbers, $\vocabattributes$ is typically much smaller. Therefore, we can effectively learn representations for entities given their set of attributes, from our set of example pairs.

\paragraph{Edge Labels}\label{edge_labels}

In addition to $\mathbf{x}$ and $\mathbf{e}$ for a particular example, we also consider the $(|\mathbf{x}|+|\mathbf{e}|)^2$ pairwise relations between all tokens and entity candidates, represented as edge labels.

The edge label between tokens $x_i$ and $x_j$ corresponds to the relative sequential position, $j - i$, of the tokens, clipped to within some range.

The edge label between token $x_i$ and entity $e_j$, and vice versa, corresponds to whether $x_i$ is within the span of the entity candidate $e_j$, or not.

The edge label between entities $e_i$ and $e_j$ captures the relationship between the entities. These edge labels can have domain-specific interpretations, such as relations in a knowledge base, or any other type of entity interaction features. For tasks where this information is not available or useful, a single generic label between entity candidates can be used.

\begin{figure*}[t!]
\begin{center}
\includegraphics[width=0.8\textwidth]{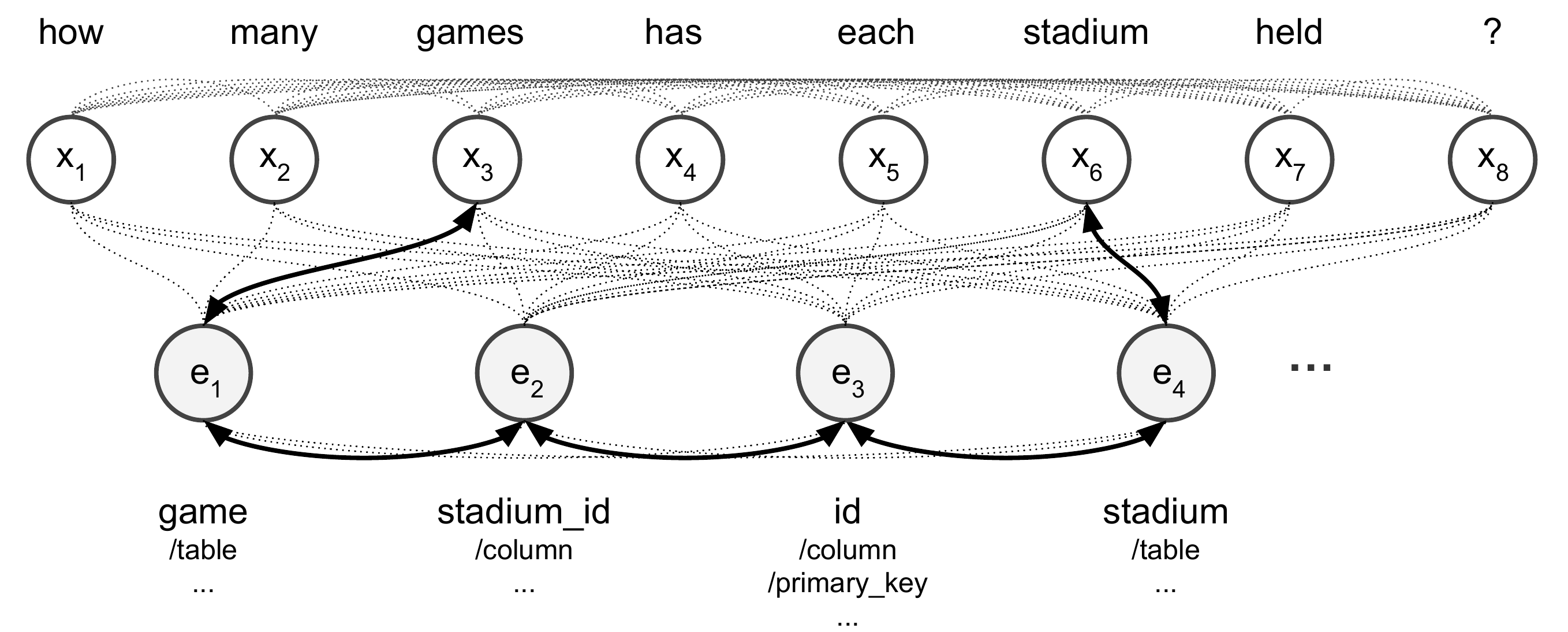}
\caption{We use an example from \spider to illustrate the model inputs: tokens from the given utterance, $\mathbf{x}$, a set of potentially relevant entities, $\mathbf{e}$, and their relations. We selected two edge label types to highlight: edges denoting that an entity spans a token, and edges between entities that, for \spider, indicate a foreign key relationship between columns, or an ownership relationship between columns and tables.}
\label{fig:example_input}
\end{center}
\end{figure*}

\paragraph{Output}

We consider the logical form, $\mathbf{y}$, to be a linear sequence~\cite{vinyals2015grammar}. We tokenize based on the syntax of each domain. 
Our formulation allows each element of $\mathbf{y}$ to be either an element of the output vocabulary, $\vocabout$, or an entity copied from the set of entity candidates $\mathbf{e}$. Therefore, $y_i \in \vocabout \cup \vocabentities$. Some experiments in \S\ref{copying_tokens} also allow elements of $\mathbf{y}$ to be tokens $\in \vocabin$ from $\mathbf{x}$ that are copied from the input.

\section{Model Architecture}

Our model architecture is based on the Transformer~\cite{vaswani2017attention}, with the self-attention sub-layer extended to incorporate relations between input elements, and the decoder extended with a copy mechanism.

\subsection{GNN Sub-layer}

We extend the Transformer's self-attention mechanism to form a Graph Neural Network (GNN) sub-layer that incorporates a fully connected, directed graph with edge labels.

The sub-layer maps an ordered sequence of node representations, $\mathbf{u} = (u_1, \ldots, u_{|\mathbf{u}|})$, to a new sequence of node representations, $\mathbf{u}^\prime = (u_1^\prime, \ldots, u_{|\mathbf{u}|}^\prime)$, where each node is represented $\in \R^d$. We use $r_{ij}$ to denote the edge label corresponding to $u_i$ and $u_j$.

We implement this sub-layer in terms of a function $f(m, l)$ over a node representation $m \in \R^d$ and an edge label $l$ that computes a vector representation in $\R^{d^\prime}$. We use $n_{heads}$ parallel attention heads, with $d^\prime = d / n_{heads}$. For each head $k$, the new representation for the node $u_i$ is computed by 

\begin{equation}\label{attention}
u^{k\prime}_i = \sum_{j=1}^{|\mathbf{u}|}
\alpha_{ij} f(u_j, r_{ij}),
\end{equation}
where each coefficient $\alpha_{ij}$ is a softmax over the scaled dot products $s_{ij}$,

\begin{equation}
s_{ij} = \frac{(\mathbf{W}^{q}u_i)^\intercal f(u_j, r_{ij})}{\sqrt{d^\prime}},
\end{equation}
and $\mathbf{W}^q$ is a learned matrix. Finally, we concatenate representations from each head, 

\begin{equation}
u_i^\prime = \mathbf{W}^h \left[ u_i^{1\prime} ~ | ~\cdots~ | ~ u_i^{n_{heads}\prime} \right],
\end{equation}
where $\mathbf{W}^h$ is another learned matrix and  $[~\cdots~]$ denotes concatenation.

If we implement $f$ as,
\begin{equation}
f(m, l) = \mathbf{W}^rm,
\end{equation}
where $\mathbf{W}^r \in \R^{d^\prime \times d}$ is a learned matrix, then the sub-layer would be effectively identical to self-attention as initially proposed in the Transformer~\cite{vaswani2017attention}.

We focus on two alternative formulations of $f$ that represent edge labels as learned matrices and learned vectors. 

\begin{figure*}[th!]
\begin{center}
\includegraphics[width=0.9\textwidth]{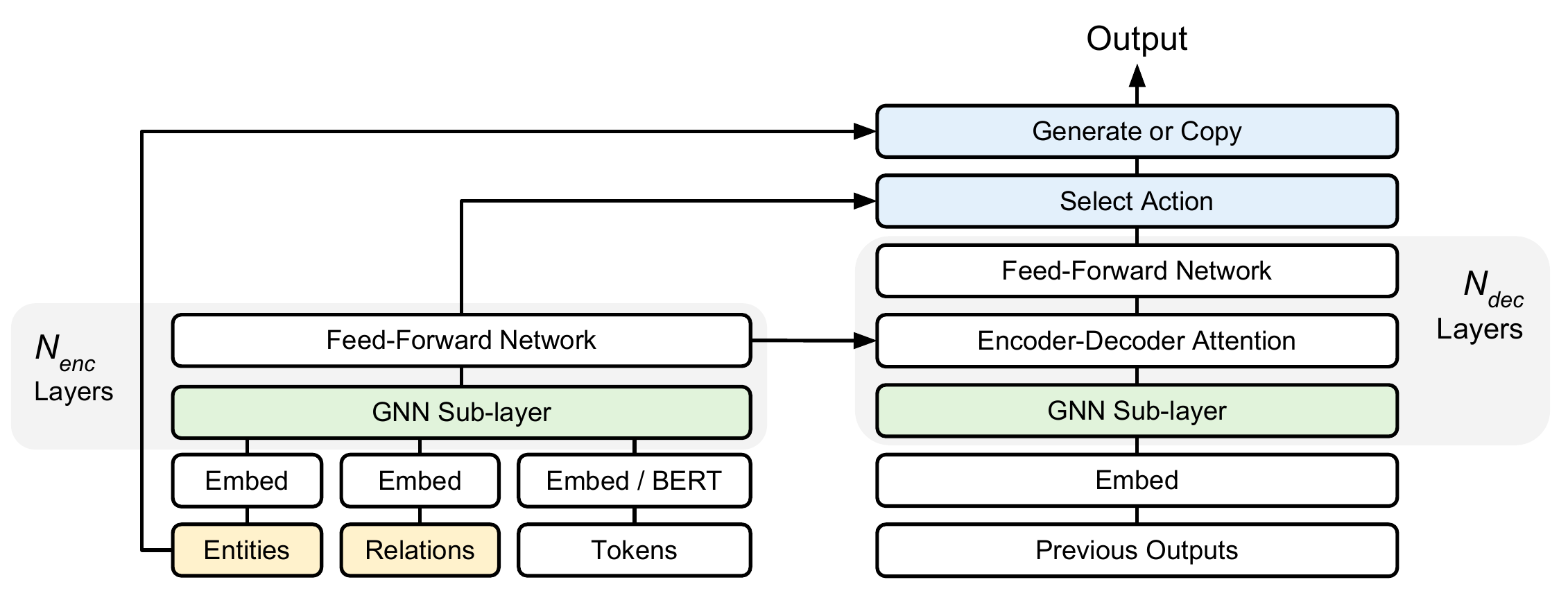}
\caption{Our model architecture is based on the Transformer~\cite{vaswani2017attention}, with two modifications. First, the self-attention sub-layer has been extended to be a GNN that incorporates edge representations. In the encoder, the GNN sub-layer is conditioned on tokens, entities, and their relations. Second, the decoder has been extended to include a copy mechanism~\cite{vinyals2015pointer}. We can optionally incorporate a pre-trained model such as BERT to generate contextual token representations.}
\label{fig:model_diagram}
\end{center}
\end{figure*}

\paragraph{Edge Matrices}
The first formulation represents edge labels as linear transformations, a common parameterization for GNNs~\cite{li2016gated},
\begin{equation}\label{edge_matrix}
f(m, l) = \mathbf{W}^{l}m,
\end{equation}
where $\mathbf{W}^l \in \R^{d^\prime \times d}$ is a learned embedding matrix per edge label. 

\paragraph{Edge Vectors}
The second formulation represents edge labels as additive vectors using the same formulation as ~\newcite{shaw2018self},
\begin{equation}\label{edge_vector}
f(m, l) = \mathbf{W}^rm+\mathbf{w}^{l},
\end{equation}
where $\mathbf{W}^r \in \R^{d^\prime \times d}$ is a learned matrix shared by all edge labels, and $w^{l} \in \R^{d}$ is a learned embedding vector per edge label $l$.

\subsection{Encoder}

\paragraph{Input Representations}
Before the initial encoder layer, tokens are mapped to initial representations using either a learned embedding table for $\vocabin$, or the output of a pre-trained BERT~\cite{devlin2018bert} encoder. Entity candidates are mapped to initial representations using the mean of the embeddings for each of the entity's attributes, based on a learned embedding table for $\vocabattributes$. We also concatenate an embedding representing the node type, token or entity, to each input representation.

We assume some arbitrary ordering for entity candidates, generating a combined sequence of initial node representations for tokens and entities. We have edge labels between every pair of nodes as described in \S~\ref{edge_labels}.

\paragraph{Encoder Layers}

Our encoder layers are essentially identical to the Transformer, except with the proposed extension to self-attention to incorporate edge labels. Therefore, each encoder layer consists of two sub-layers. The first is the GNN sub-layer, which yields new sets of token and entity representations. The second sub-layer is an element-wise feed-forward network. Each sub-layer is followed by a residual connection and layer normalization~\cite{ba2016layer}. We stack $N_{enc}$ encoder layers, yielding a final set of token representations, $\mathbf{w_x}^{(N_{enc})}$, and entity representations, $\mathbf{w_e}^{(N_{enc})}$. 

\subsection{Decoder}\label{decoder}

The decoder auto-regressively generates output symbols, $y_1, \ldots, y_{|\mathbf{y}|}$. It is similarly based on the Transformer~\cite{vaswani2017attention}, with the self-attention sub-layer replaced by the GNN sub-layer. Decoder edge labels are based only on the relative timesteps of the previous outputs. The encoder-decoder attention layer considers both encoder outputs $\mathbf{w_x}^{(N_{enc})}$ and $\mathbf{w_e}^{(N_{enc})}$, jointly normalizing attention weights over tokens and entity candidates. We stack $N_{dec}$ decoder layers to produce an output vector representation at each output step, $z_j \in \R^{d_z}$, for $j \in \{1, \ldots, |\mathbf{y}|\}$.

We allow the decoder to copy tokens or entity candidates from the input, effectively combining a Pointer Network~\cite{vinyals2015pointer} with a standard softmax output layer for selecting symbols from an output vocabulary~\cite{gu2016incorporating, gulcehre2016pointing, jia2016data}. We define a latent action at each output step, $a_j$ for $j \in \{1,\ldots,|\mathbf{y}|\}$, using similar notation as Jia et al.~\shortcite{jia2016data}. We normalize action probabilities with a softmax over all possible actions.

\paragraph{Generating Symbols}

We can generate a symbol, denoted $\texttt{Generate}[i]$,
\begin{align}
\begin{split}
  P(a_j = &\texttt{Generate}[i] \mid \mathbf{x}, \mathbf{y}_{1:j-1}) \propto \\
                                & \exp(z_j^\intercal w^{out}_i), \\
\end{split}
\end{align}
where $w^{out}_i$ is a learned embedding vector for the element $\in \vocabout$ with index $i$. If $a_j = \texttt{Generate}[i]$, then $y_j$ will be the element  $\in \vocabout$ with index $i$.

\paragraph{Copying Entities}
We can also copy an entity candidate, denoted $\texttt{CopyEntity}[i]$,
\begin{align}
\begin{split}
  P(a_j & = \texttt{CopyEntity}[i] \mid \mathbf{x}, \mathbf{y}_{1:j-1}) \propto \\
                              & \exp((z_j\mathbf{W}^e)^\intercal w^{(N_{enc})}_{e_i}), \\
\end{split}
\end{align}
where $\mathbf{W}^e$ is a learned matrix, and $i \in \{1,\ldots,|\mathbf{e}|\}$. If $a_j = \texttt{CopyEntity}[i]$, then $y_j = e_i$. 

\section{Related Work}

Various approaches to learning semantic parsers from pairs of utterances and logical forms have been developed over the years ~\cite{tang2000automated, zettlemoyer2007online, kwiatkowski2011lexical, andreas2013semantic}. More recently, encoder-decoder architectures have been applied with strong results ~\cite{dong2016language, jia2016data}.

Even for tasks with relatively small domains of entities, such as \geo and \atis, it has been shown that some special consideration of entities within an encoder-decoder architecture is important to improve generalization. This has included extending decoders with copy mechanisms ~\cite{jia2016data} and/or identifying entities in the input as a pre-processing step ~\cite{dong2016language}.

Other work has considered open domain tasks, such as \webquestions~\cite{yih2016value}. Recent approaches have typically relied on a separate entity linking model, such as S-MART~\cite{yang2015smart}, to provide a single disambiguated set of entities to consider. In principle, a learned entity linker could also serve as an entity candidate generator within our framework, although we do not explore such tasks in this work.

Considerable recent work has focused on constrained decoding of various forms within an encoder-decoder architecture to leverage the known structure of the logical forms. This has led to approaches that leverage this structure during decoding, such as using tree decoders \cite{dong2016language, alvarez2017tree} or other mechanisms \cite{dong2018coarse, goldman2017weakly}. Other approaches use grammar rules to constrain decoding \cite{xiao2016sequence,yin2017syntactic,krishnamurthy2017neural,yu2018syntaxsqlnet}. We leave investigation of such decoder constraints to future work.

Many formulations of Graph Neural Networks (GNNs) that propagate information over local neighborhoods have recently been proposed~\cite{li2016gated,kipf2016semi,gilmer2017neural,velivckovic2017graph}.
Recent work has often focused on large graphs~\cite{hamilton2017inductive} and effectively propagating information over multiple graph steps~\cite{xu2018representation}. The graphs we consider are relatively small and are fully-connected, avoiding some of the challenges posed by learning representations for large, sparsely connected graphs.

Other recent work related to ours has considered GNNs for natural language tasks, such as combining structured and unstructured data for question answering~\cite{sun2018open}, or for representing dependencies in tasks such as AMR parsing and machine translation~\cite{beck2018graph, bastings2017graph}. The approach of ~\newcite{krishnamurthy2017neural} similarly considers ambiguous entity mentions jointly with query tokens for semantic parsing, although does not directly consider a GNN.

Previous work has interpreted the Transformer's self-attention mechanism as a GNN~\cite{velivckovic2017graph, battaglia2018relational}, and extended it to consider relative positions as edge representations~\cite{shaw2018self}. Previous work has also similarly represented edge labels as vectors, as opposed to matrices, in order to avoid over-parameterizing the model~\cite{marcheggiani2017encoding}.

\section{Experiments}

\subsection{Semantic Parsing Datasets}

We consider three semantic parsing datasets, with examples given in Table~\ref{semantic_parsing_tasks}.

\paragraph{\geo} The GeoQuery dataset consists of natural language questions about US geography along with corresponding logical forms~\cite{zelle1996learning}. We follow the convention of ~\newcite{zettlemoyer2005learning} and use 600 training examples and 280 test examples. We use logical forms based on Functional Query Language (FunQL)~\cite{kate2005learning}. 

\paragraph{\atis} The Air Travel Information System (\atis) dataset consists of natural language queries about travel planning~\cite{dahl1994expanding}. We follow ~\newcite{zettlemoyer2007online} and use 4473 training examples, 448 test examples, and represent the logical forms as lambda expressions.

\paragraph{\spider} This is a large-scale text-to-SQL dataset that consists of 10,181 questions and 5,693 unique complex SQL queries across 200 database tables spanning 138 domains~\cite{yu2018spider}. We use the standard training set of 8,659 training example and development set of 1,034 examples, split across different tables.

\subsection{Experimental Setup}

\paragraph{Model Configuration} 

We configured hyperparameters based on performance on the validation set for each task, if provided, otherwise cross-validated on the training set.

For the encoder and decoder, we selected the number of layers from $\{1,2,3,4\}$ and embedding and hidden dimensions from $\{64,128,256\}$, setting the feed forward layer hidden dimensions $4\times$ higher. 
We employed dropout at training time with $P_{dropout}$ selected from $\{0.1, 0.2, 0.3, 0.4, 0.5, 0.6\}$. We used $8$ attention heads for each task. We used a clipping distance of $8$ for relative position representations~\cite{shaw2018self}.

We used the Adam optimizer ~\cite{kingma2014adam} with $\beta_1=0.9$, $\beta_2=0.98$, and $\epsilon = 10^{-9}$, and tuned the learning rate for each task. We used the same warmup and decay strategy for learning rate as Vaswani et
al.~\shortcite{vaswani2017attention}, selecting a number of warmup steps up to a maximum of 3000. Early stopping was used to determine the total training steps for each task. We used the final checkpoint for evaluation. We batched training examples together, and selected batch size from $\{32, 64, 128, 256, 512\}$. During training we used masked self-attention ~\cite{vaswani2017attention} to enable parallel decoding of output sequences. For evaluation, we used greedy search.

We used a simple strategy of splitting each input utterance on spaces to generate a sequence of tokens. We mapped any token that didn't occur at least 2 times in the training dataset to a special out-of-vocabulary token. For experiments that used BERT, we instead used the same wordpiece~\cite{wu2016google} tokenization as used for pre-training.

\paragraph{BERT}

For some of our experiments, we evaluated incorporating a pre-trained BERT~\cite{devlin2018bert} encoder by effectively using the output of the BERT encoder in place of a learned token embedding table. We then continue to use graph encoder and decoder layers with randomly initialized parameters in addition to BERT, so there are many parameters that are not pre-trained. The additional encoder layers are still necessary to condition on entities and relations. 

We achieved best results by freezing the pre-trained parameters for an initial number of steps, and then jointly fine-tuning all parameters, similar to existing approaches for gradual unfreezing~\cite{howard2018universal}. When unfreezing the pre-trained parameters, we restart the learning rate schedule. We found this to perform better than keeping pre-trained parameters either entirely frozen or entirely unfrozen during fine-tuning.

We used \bertlarge~\cite{devlin2018bert}, which has 24 layers. For fine tuning we used the same Adam optimizer with weight decay and learning rate decay as used for BERT pre-training. We reduced batch sizes to accommodate the significantly larger model size, and tuned learning rate, warm up steps, and number of frozen steps for pre-trained parameters.

\paragraph{Entity Candidate Generator}\label{entity_candidate_generator}

We use an entity candidate generator that, given $\mathbf{x}$, can retrieve a set of potentially relevant entities, $\mathbf{e}$, for the given domain. Although all generators share a common interface, their implementation varies across tasks. 

For \geo and \atis we use a lexicon of entity aliases in the dataset and attempt to match with ngrams in the query. Each entity has a single attribute corresponding to the entity's type. We used binary valued relations between entity candidates based on whether entity candidate spans overlap, but experiments did not show significant improvements from incorporating these relations.

For \spider, we generalize our notion of entities to include tables and table columns. 
We include all relevant tables and columns as entity candidates, but make use of Levenshtein distance between query ngrams and table and column names to determine edges between tokens and entity candidates. We use attributes based on the types and names of tables and columns. Edges between entity candidates capture relations between columns and the table they belong to, and foreign key relations.

For \geo, \atis, and \spider, this leads to 19.5\%, 32.7\%, and 74.6\% of examples containing at least one span associated with multiple entity candidates, respectively, indicating some entity ambiguity.

Further details on how entity candidate generators were constructed are provided in \S~\ref{entity_candidate_generator_details}.

\begin{table*}[!ht]
\begin{center}
\begin{minipage}{.55\linewidth}
\scalebox{0.9}{
\begin{tabular}{lrrr}
\toprule
\bf Method & \bf \geo & \bf \atis  \\
\midrule
~\newcite{kwiatkowski2013scaling} & 89.0 & ---  \\
~\newcite{liang2013learning} & 87.9 & --- \\
~\newcite{wang2014morpho} & --- & \bf{91.3} \\
~\newcite{zhao2015type} & 88.9 & 84.2 \\
~\newcite{jia2016data} & 89.3 & 83.3  \\
~~~~ $-$ data augmentation & 85.0 & 76.3 \\
~\newcite{dong2016language} $\dagger$ & 87.1 & 84.6  \\
~\newcite{rabinovich2017abstract} $\dagger$ & 87.1 & 85.9  \\
~\newcite{dong2018coarse} $\dagger$ & 88.2 & 87.7  \\
\midrule
\bf \emph{Ours} & & & \\
\midrule
GNN w/ edge matrices & 82.5 & 84.6  \\
GNN w/ edge vectors & 89.3 & 87.1 \\
GNN w/ edge vectors + BERT  & \bf{92.5} & 89.7 \\
\bottomrule
\end{tabular}
}
\end{minipage}
\begin{minipage}{.4\linewidth}
\scalebox{0.9}{
\begin{tabular}{lr}
\toprule
\bf Method & \bf \spider \\
\midrule
~\newcite{xu2017sqlnet}   & 10.9 \\
~\newcite{yu2018typesql}  &  8.0 \\
~\newcite{yu2018syntaxsqlnet} & 24.8 \\
~~~~ $-$ data augmentation & 18.9 \\
\midrule
\bf \emph{Ours} & \\ 
\midrule
GNN w/ edge matrices & 29.3 \\
GNN w/ edge vectors & \bf 32.1 \\
GNN w/ edge vectors + BERT & 23.5 \\ 
\bottomrule
\end{tabular}
}
\end{minipage}
\end{center}

\caption{We report accuracies on \geo, \atis, and \spider for various implementations of our GNN sub-layer. For \geo and \atis, we use $\dagger$ to denote neural approaches that disambiguate and replace entities in the utterance as a pre-processing step. For \spider, the evaluation set consists of examples for databases unseen during training. }

\label{experiment_results}
\end{table*}

\paragraph{Output Sequences}

We pre-processed output sequences to identify entity argument values, and replaced those elements with references to entity candidates in the input. In cases where our entity candidate generator did not retrieve an entity that was used as an argument, we dropped the example from the training data set or considered it incorrect if in the test set. 

\paragraph{Evaluation}

To evaluate accuracy, we use exact match accuracy relative to gold logical forms. For \geo we directly compare output symbols. For \atis, we compare normalized logical forms using canonical variable naming and sorting for un-ordered arguments~\cite{jia2016data}. For \spider we use the provided evaluation script, which decomposes each SQL query and conducts set comparison within each clause without values. All accuracies are reported on the test set, except for \spider where we report and compare accuracies on the development set.

\paragraph{Copying Tokens}\label{copying_tokens}
To better understand the effect of conditioning on entities and their relations, we also conducted experiments that considered an alternative method for selecting and disambiguating entities similar to Jia et al.~\shortcite{jia2016data}. In this approach we use our model's copy mechanism to copy tokens corresponding to the surface forms of entity arguments, rather than copying entities directly.
\begin{align}
\begin{split}
  P(a_j & = \texttt{CopyToken}[i] \mid \mathbf{x}, \mathbf{y}_{1:j-1}) \propto \\
                             & \exp((z_j\mathbf{W}^x)^\intercal w^{(N_{enc})}_{x_i}), \\
\end{split}
\end{align}
where $\mathbf{W}^x$ is a learned matrix, and where $i \in \{1, \ldots, |\mathbf{x}|\}$ refers to the index of token $x_i \in \vocabin$. If $a_j = \texttt{CopyToken}[i]$, then $y_j = x_i$.

This allows us to ablate entity information in the input while still generating logical forms. When copying tokens, the decoder determines the type of the entity using an additional output symbol. For \geo, the actual entity can then be identified as a post-processing step, as a type and surface form is sufficient. For other tasks this could require a more complicated post-processing step to disambiguate entities given a surface form and type.

\begin{table}[h!]
\begin{center}
\scalebox{0.9}{
\begin{tabular}{lr}
\toprule
\bf Method & \bf \geo \\
\midrule
 \bf \emph{Copying Entities} & \\[3pt]
 ~~~~ GNN w/ edge vectors + BERT & \bf 92.5 \\
 ~~~~ GNN w/ edge vectors & 89.3 \\
 \midrule
 \bf \emph{Copying Tokens} & \\[3pt]
~~~~GNN w/ edge vectors & 87.9 \\
~~~~ ~~~~ $-$ entity candidates, $\mathbf{e}$ & 84.3  \\
~~~~BERT & 89.6 \\
\bottomrule
\end{tabular}
}
\end{center}
\caption{Experimental results for copying tokens instead of entities when decoding, with and without conditioning on the set of entity candidates, $\mathbf{e}$.}
\label{copy_tokens_results}
\end{table}

\subsection{Results and Analysis}

Accuracies on \geo, \atis, and \spider are shown in Table~\ref{experiment_results}.

\paragraph{\geo and \atis} Without pre-training, and despite adding a bit of entity ambiguity, we achieve similar results to other recent approaches that disambiguate and replace entities in the utterance as a pre-processing step during both training and evaluating~\cite{dong2016language, dong2018coarse}. When incorporating BERT, we increase absolute accuracies over ~\newcite{dong2018coarse} on \geo and \atis by 3.2\% and 2.0\%, respectively. Notably, they also present techniques and results that leverage constrained decoding, which our approach would also likely further benefit from.

For \geo, we find that when ablating all entity information in our model and copying tokens instead of entities, we achieve similar results as Jia and Liang~\shortcite{jia2016data} when also ablating their data augmentation method, as shown in Table~\ref{copy_tokens_results}. This is expected, since when ablating entities completely, our architecture essentially reduces to the same sequence-to-sequence task setup. These results demonstrate the impact of conditioning on the entity candidates, as it improves performance even on the token copying setup. It appears that leveraging BERT can partly compensate for not conditioning on entity candidates, but combining BERT with our GNN approach and copying entities achieves 2.9\% higher accuracy than using only a BERT encoder and copying tokens.

For \atis, our results are outperformed by ~\newcite{wang2014morpho} by 1.6\%. Their approach uses hand-engineered templates to build a CCG lexicon. Some of these templates attempt to handle the specific types of ungrammatical utterances in the ATIS task.

\paragraph{\spider} For \spider, a relatively new dataset, there is less prior work. Competitive approaches have been specific to the text-to-SQL task ~\cite{xu2017sqlnet, yu2018typesql, yu2018syntaxsqlnet}, incorporating task-specific methods to condition on table and column information, and incorporating SQL-specific structure when decoding. Our approach improves absolute accuracy by +7.3\% relative to Yu et al.~\shortcite{yu2018syntaxsqlnet} without using any pre-trained language representations, or constrained decoding. Our approach could also likely benefit from some of the other aspects of Yu et al.~\shortcite{yu2018syntaxsqlnet} such as more structured decoding, data augmentation, and using pre-trained representations (they use GloVe~\cite{pennington2014glove}) for tokens, columns, and tables.

Our results were surprisingly worse when attempting to incorporate BERT. Of course, successfully incorporating pre-trained representations is not always straightforward. In general, we found using BERT within our architecture to be sensitive to learning rates and learning rate schedules. Notably, the evaluation setup for \spider is very different than training, as examples are for tables unseen during training. Models may not generalize well to unseen tables and columns. It's likely that successfully incorporating BERT for \spider would require careful tuning of hyperparameters specifically for the database split configuration.

\paragraph{Entity Spans and Relations} 

Ablating span relations between entities and tokens for \geo and \atis is shown in Table~\ref{relation_ablation_geo_atis}. The impact is more significant for \atis, which contains many queries with multiple entities of the same type, such as \emph{nonstop flights seattle to boston} where disambiguating the origin and destination entities requires knowledge of which tokens they are associated with, given that we represent entities based only on their types for these tasks. We leave for future work consideration of edges between entity candidates that incorporate relevant domain knowledge for these tasks.

\begin{table}[h!]
\begin{center}
\scalebox{0.9}{
\begin{tabular}{lrr}
\toprule
\bf Edge Ablations & \bf \geo & \bf \atis \\
\midrule
GNN w/ edge vectors & 89.3 & 87.1 \\
~~~~ $-$ entity span edges & 88.6 & 34.2 \\
\bottomrule
\end{tabular}
}
\end{center}
\caption{Results for ablating information about entity candidate spans for \geo and \atis.}
\label{relation_ablation_geo_atis}
\end{table}

For \spider, results ablating relations between entities and tokens, and relations between entities, are shown in Table~\ref{relation_ablation_spider}. This demonstrates the importance of entity relations, as they include useful information for disambiguating entities such as which columns belong to which tables.

\begin{table}[h!]
\begin{center}
\scalebox{0.9}{
\begin{tabular}{lr}
\toprule
\bf Edge Ablations & \bf \spider \\
\midrule
GNN w/ edge vectors & 32.1 \\
~~~~ $-$ entity span edges & 27.8 \\
~~~~ $-$ entity relation edges & 26.3 \\
\bottomrule
\end{tabular}
}
\end{center}
\caption{Results for ablating information about relations between entity candidates and tokens for \spider.}
\label{relation_ablation_spider}
\end{table}

\paragraph{Edge Representations} Using additive edge vectors outperforms using learned edge matrix transformations for implementing $f$, across all tasks. While the vector formulation is less expressive, it also introduces far fewer parameters per edge type, which can be an important consideration given that our graph contains many similar edge labels, such as those representing similar relative positions between tokens. We leave further exploration of more expressive edge representations to future work. Another direction to explore is a heterogeneous formulation of the GNN sub-layer, that employs different formulations for different subsets of nodes, e.g. for tokens and entities.

\section{Conclusions}

We have presented an architecture for semantic parsing that uses a Graph Neural Network (GNN) to condition on a graph of tokens, entities, and their relations. Experimental results have demonstrated that this approach can achieve competitive results across a diverse set of tasks, while also providing a conceptually simple way to incorporate entities and their relations during parsing.

For future direction, we are interested in exploring constrained decoding, better incorporating pre-trained language representations within our architecture, conditioning on additional relations between entities, and different GNN formulations.

More broadly, we have presented a flexible approach for conditioning on available knowledge in the form of entities and their relations, and demonstrated its effectiveness for semantic parsing.

\section*{Acknowledgments}

We would like to thank Karl Pichotta, Zuyao Li, Tom Kwiatkowski, and Dipanjan Das for helpful discussions. Thanks also to Ming-Wei Chang and Kristina Toutanova for their comments, and to all who provided feedback in draft reading sessions. Finally, we are grateful to the anonymous reviewers for their useful feedback.

\bibliographystyle{acl_natbib}
\bibliography{paper}

\clearpage
\appendix

\section{Supplemental Material}
\label{sec:supplemental}
\subsection{Entity Candidate Generator Details}\label{entity_candidate_generator_details}

In this section we provide details of how we constructed entity candidate generators for each task.

\paragraph{\geo}
The annotator was constructed from the geobase database, which provides a list of geographical facts. For each entry in the database, we extracted the name as the entity alias and the type (e.g., ``state'', ``city'') as its attribute. Since not all cities used in the \geo query set are listed as explicit entries, we also used cities in the state entries. Finally, geobase has no entries around countries, so we added relevant aliases for ``USA'' with a special ``country'' attribute. There was 1 example where an entity in the logical form did not appear in the input, leading to the example being dropped from the training set.

In lieu of task-specific edge relations, we used binary edge labels between entities that captured which annotations span the same tokens. However, experiments demonstrated that these edges did not significantly affect performance. We leave consideration of other types of entity relations for these tasks to future work.

\paragraph{\atis}
We constructed a lexicon mapping natural language entity aliases in the dataset (e.g., ``newark international'', ``5pm'') to unique entity identifiers (e.g. ``ewr:ap'', ``1700:ti''). For \atis, this lexicon required some manual construction. Each entity identifier has a two-letter suffix (e.g., ``ap'', ``ti'') that maps it to a single attribute (e.g., ``airport'', ``time''). We allowed overlapping entity mentions when the entities referred to different entity identifiers. For instance, in the query span ``atlanta airport'', we include both the city of Atlanta and the Atlanta airport.

Notably there were 9 examples where one of the entities used as an argument in the logical form did not have a corresponding mention in the input utterance. From manual inspection, many of the dropped examples appear to have incorrectly annotated logical forms. These examples were dropped from training set or marked as incorrect if they appeared in the test set.

We use the same binary edge labels between entities as for \geo.

\paragraph{\spider}

For \spider we generalize our notion of entities to consider tables and columns as entities. We attempt to determine spans for each table and column by computing normalized Levenshtein distance between table and column names and unigrams or bigrams in the utterance. The best alignment having a score $>0.75$ is selected, and we use these generated alignments to populate the edges between tokens and entity candidates.

We generate a set of attributes for the table based on unigrams in the table name, and an attribute to identify the entity as a table. Likewise, for columns, we generate a set of attributes based on unigrams in the column name as well as an attribute to identify the value type of the column. We also include attributes indicating whether an alignment was found between the entity and the input text.

We include 3 task-specific edge label types between entity candidates to denote bi-directional relations between column entities and the table entity they belong to, and to denote the presence of a foreign key relationship\footnote{Due to a bug in our pre-processing code for the SPIDER dataset, only at most one foreign key edge relation per example was included for the experiments and results presented in this paper. However, resolving this issue and including all foreign key relations does not improve the overall accuracy results, an interesting observation for future consideration. Thanks to Alane Suhr for identifying this issue.} between columns.

\end{document}